\documentclass[hidelinks,12pt]{iopart}
\usepackage[T1]{fontenc}
\usepackage[utf8]{inputenc}
\usepackage{lmodern}
\usepackage[english]{babel}
\expandafter\let\csname equation*\endcsname\relax % for iopart / amsmath error
\expandafter\let\csname endequation*\endcsname\relax 
\usepackage{amsmath}
\usepackage{amssymb}
\usepackage{acronym}
\usepackage{graphicx}
\usepackage{xcolor}
\usepackage{url}
\usepackage{multirow}
\usepackage{hhline}
\usepackage{hyperref}
\usepackage{cleveref}

\usepackage{subcaption}
\crefname{section}{Sec.}{Secs.}
\crefname{figure}{Fig.}{Figs.}

\definecolor{colTextGT}{RGB}{50, 140, 54}
\definecolor{colTextP}{RGB}{149,30,26}

\makeatletter
\long\def\@makefntext#1{\parindent 1em\noindent 
 \makebox[1em][l]{\footnotesize\rm$\m@th{\arabic{footnote}}$}%
 \footnotesize\rm #1}
\def\@makefnmark{\hbox{$^{\arabic{footnote}}\m@th$}}
\def\@thefnmark{\arabic{footnote}}
\makeatother

\begin{document}
\setcounter{footnote}{0}

%\title{Transfer Learning pipeline for Few-Shot Object Detection with diffusion models}
%or

%\title{Generative Transfer Learning for Cross-Domain Object Detection with Minimal Real Data}
%or

% \title{Transfer learning for object detection with diffusion generative models on limited datasets}

\title{Transfer learning with generative models for object detection on limited datasets}

\author{M. Paiano$^1$, S. Martina$^{2,3}$, C. Giannelli$^1$, F. Caruso$^{2,3,4}$}

\address{$^1$Department of Mathematics and Computer Science,
University of Florence, Viale  Morgagni 67/a, I-50134 Florence, Italy}
\address{$^2$Department of Physics and Astronomy, University of Florence, Via Sansone 1, I-50019 Sesto Fiorentino, Italy}
\address{$^3$European Laboratory for Non-Linear Spectroscopy (LENS), University of Florence, Via Nello Carrara 1, Sesto Fiorentino, I-50019, Italy}
\address{$^4$Istituto Nazionale di Ottica, Consiglio Nazionale delle Ricerche (CNR-INO), I-50019 Sesto Fiorentino, Italy}

\ead{matteo.paiano@unifi.it, stefano.martina@unifi.it, carlotta.giannelli@unifi.it, filippo.caruso@unifi.it}

\vspace{10pt}
% \begin{indented}
% \item[]August 2017
% \end{indented}

\begin{abstract}
The availability of data is limited in some fields, especially for object detection tasks, where it is necessary to have correctly labeled bounding boxes around each object. A notable example of such data scarcity is found in the domain of marine biology, where it is useful to develop methods to automatically detect submarine species for environmental monitoring. To address this data limitation, the state-of-the-art machine learning strategies employ two main approaches. The first involves pretraining models on existing datasets before generalizing to the specific domain of interest. The second strategy is to create synthetic datasets specifically tailored to the target domain using methods like copy-paste techniques or ad-hoc simulators. The first strategy often faces a significant domain shift, while the second demands custom solutions crafted for the specific task. 
In response to these challenges, here we propose a transfer learning framework that is valid for a generic scenario. In this framework, generated images help to improve the performances of an object detector in a few-real data regime. This is achieved through a diffusion-based generative model that was pretrained on large generic datasets. With respect to the state-of-the-art, we find that it is not necessary to fine tune the generative model on the specific domain of interest. We believe that this is an important advance because it mitigates the labor-intensive task of manual labeling the images in object detection tasks.
We validate our approach focusing on fishes in an underwater environment, and on the more common domain of cars in an urban setting. Our method achieves detection performance comparable to models trained on thousands of images, using only a few hundreds of input data. Our results pave the way for new generative AI-based protocols for machine learning applications in various domains, for instance ranging from geophysics to biology and medicine.
\end{abstract}
\vspace{2pc}
\noindent{\it Keywords}: Object Detection, Transfer Learning, Generative AI, Diffusion Models, Deep Learning
%\maketitle

\acrodef{ai}[AI]{Artificial Intelligence}
\acrodef{ann}[ANN]{Artificial Neural Network}
\acrodef{ml}[ML]{Machine Learning}
\acrodef{gan}[GAN]{Generative Adversarial Network}
\acrodef{ddpm}[DDPM]{Denoising Diffusion Probabilistic Model}
\acrodef{sgm}[SGM]{Score-based Generative Model}
\acrodef{sde}[SDE]{Stochastic Differential Equation}
\acrodef{l2i}[L2I]{Layout-to-Image}
\acrodef{map}[mAP]{mean Average Precision}
\acrodef{cnn}[CNN]{Convolutional Neural Network}
\acrodef{rpn}[RPN]{Region Proposal Network}
\acrodef{fid}[FID]{Frechet Inception Distance}
\acrodef{iou}[IoU]{Intersection over Union}
\acrodef{sgd}[SGD]{Stochastic Gradient Descent}
\acrodef{fpn}[FPN]{Feature Pyramid Network}
\acrodef{fcos}[FCOS]{Fully Convolutional One-Stage object detector}

\section{Introduction}
\label{sec:introduction}
\ac{ml} is the area of artificial intelligence that involves the development of models that are trained on data to perform specific jobs.
In the last few years, we are witnessing a growing interest in generative models as powerful tools in the context of \ac{ml}. 
Generative models are capable of estimating the unknown probability distribution underlying the training dataset, in order to generate samples from such distribution.
Some of the research works in this field~\cite{harshvardhan2020generative} include Gaussian Mixture Models, Hidden Markov Models, Latent Dirichlet Allocation, Boltzmann Machines and, lately, Variational Autoencoders and \acp{gan}~\cite{ruthotto2021generative}.
More recently, \emph{diffusion} models emerged as a new family of generative models~\cite{yang2023diffusion}, overcoming \acp{gan} as the most adopted and powerful generative models in image generation tasks~\cite{croitoru2023diffusion}. They are designed according to three main formulations: \acp{ddpm}, \acp{sgm}, or \acp{sde}. In all these approaches, the main concept is that the images of the training dataset are progressively perturbed with random noise in a fixed process (called \emph{diffusion}) where all the information is destroyed, and the images at the end of the process belong to an uninformative distribution (usually a standard multivariate Gaussian distribution). Afterward, the process is reversed with a parametrized \ac{ml} model that estimates the unknown original data distribution. The trained model can then generate novel data samples belonging to an approximation of the original distribution. The most commonly adopted \acp{ddpm} use two Markov chains for the forward and backward training processes, respectively. 

Given the late increasing amount of works on generative models for image generation, we are interested in employing them in novel valuable scientific applications.
For instance, \emph{object detection} is the image processing task that consists in the identification of object instances within images~\cite{zou2023object}. One of the traditional \ac{ml} approaches is \emph{supervised learning} that requires the input data to be labelled with the desired output of the model to be trained. The model learns to extract statistical \emph{features} (patterns that capture relevant information about the content of an image) and to predict \emph{labels} for each input. For \ac{ml} object detection, labels are in the form of coordinates of bounding boxes around each target, along with identifications of objects types called \emph{classes}. This means that for training \ac{ml} models, it is desirable to have access to large amounts of correctly labelled data, but the labelling work is burdensome, and the availability of high quality annotated datasets is scarce, especially for crowded scenarios or specialized contexts. In fact, data scarcity, along with high model complexity, contributes to \emph{overfitting}: the model learns features that are too specific to the training set, sacrificing generalization capabilities on the unseen data.
To tackle this challenge, various techniques revolve around the notion of dataset \emph{domain}, defined as the joint probability distribution of the feature-label space~\cite{farahani2021brief}. 

The most common strategy to addresses the problem of data scarcity in any \ac{ml} problem in general, and in computer vision tasks in particular, is \emph{data augmentation}, which extract more information from the training datasets and enhance their diversity~\cite{shorten2019survey}. In detail, it consists in an image manipulation process that creates new images by applying on the existing dataset operations such as color space transformations, scene cropping, noise injection or affine transformations (translations, rotations, etc\dots). 
Another way to artificially expand the size and diversity of the dataset strongly relies on domain knowledge and consists in the construction of \emph{synthetic data} that mimic the global distribution of real data instances~\cite{gaidon2016virtual}. However, such data synthesizers are typically domain-specific, hard to implement, and they require a deep knowledge of the real context captured by the images. Thus, the cost of producing photorealistic images undermines the core advantage of synthetic data, which is the availability of large labeled data sets for free.
The data augmentation techniques can be thought as an equivalent of the role of imagination in humans, where alterations of the available dataset are imagined~\cite{shorten2019survey}. In fact, imagination is shown to be important in learning tasks for humans, and forming mental images helps in memory related tasks~\cite{mguidich2023does}. In particular, for object detection tasks, there is evidence that the lack of imagination in aphantasic subjects (people unable to create mental imagery) have a negative impact on the performances of these tasks~\cite{monzel2021imagine,monzel2023s}.

Augmentations and synthesizers introduce variability, but data still belongs to the existing domain. An alternative approach is \emph{domain randomization}~\cite{tremblay2018randomization}, which change the domain fundamental distribution (e.g. by randomizing foreground textures, lighting conditions, backgrounds and occlusions). This forces the model to focus only on the essential features, by interpreting the real-world data just as another variation among the synthetic data. However, this technique still demands to be familiar with the tools necessary to implement the simulation. An easier approach is to copy and paste real images of objects onto background images~\cite{dwibedi2017cut}, but this arises the challenge for accurate segmentation of the objects to be copied.

The previous approaches rely on the assumption that both train and test datasets are drawn from the same (possibly randomized) data distribution, otherwise the model will underperform on the test set due to a phenomenon known as \emph{domain shift}. To counter this problem, the common strategy is called \emph{transfer learning}~\cite{zhuang2021transfer}. The idea is based on the generalization theory of transfer of training, studied within the framework of educational psychology~\cite{judd1915psychology}. According to this theory, engaging with a new activity by establishing connections with previous experiences enhances the learning process. 
For instance, students that studied Latin learn related languages (e.g., German) more easily~\cite{brownell1936theoretical}.
Inspired by this concept, transfer learning in \ac{ml} aims to enhance model performances by transferring knowledge from one domain (referred to as the \emph{source}) to another (the \emph{target}), thereby reducing the need for domain-specific data in the target domain. One effective approach involves training a model on a specific task, such as feature extraction from a particular dataset, and subsequently applying it to extract features in a different domain. If images from the target domain are available, another common technique initially trains the model on the source dataset and then conducts a \emph{fine-tuning} stage, i.e. a re-training of part (or all) of the pretrained model using new data from the target domain~\cite{church2021emerging}. Some examples of transfer learning include, but are not limited to, image classification in the medical field~\cite{deepak2019brain, jaiswal2021classification}, detection of gravitational waves~\cite{george2018classification} and material properties prediction~\cite{jha2019enhancing}. For an extensive review of transfer learning applications, one can consult Ref.~\cite{iman2023review}.
Transfer learning can be split into two main categories: \emph{homogeneous}, when the source and target domains share the same features space, and \emph{heterogeneous}, otherwise~\cite{day2017survey}. Within homogeneous transfer learning, a subcategory known as \emph{domain adaptation}~\cite{liu2020heterogeneous} addresses scenarios where the feature spaces exhibit different distributions and aims to minimize the gap between the two domains. An illustrative example of domain adaptation is the transfer of knowledge from real pictures to cartoon images~\cite{zheng2020cartoon}. Despite sharing the same feature spaces, the distributions of colors and shapes between the two domains differ.

In many scientific areas, detection tasks can also be useful for research activities. Some examples are the counting of microorganisms~\cite{zhang2022comprehensive}, the monitoring of animals in the wild~\cite{arteta2016counting, akccay2020automated} and plants phenotyping~\cite{gomez2021deep}.
In this work, we focus on the problem of training object detectors on domains where the availability of data is scarce. For instance, in the context of marine biology, it is crucial to be able to monitor wild fishes and, in general, to properly support environmental monitoring and protection.
For research works where this task was addressed with deep learning techniques see, e.g., Refs.~\cite{villon2018deep,marini2018tracking,crescitelli2021norfisk,marrable2022accelerating}. However, the availability of data in this field is not of great extent~\cite{ditria2021annotated,veiga2022autonomous,francescangeli2023image}. To resolve this issue, some works addressed the data scarcity with the aid of simulators to generate synthetic data~\cite{ishiwaka2021foids,pedersen2023brackishmot}; other solutions have been found in the context of transfer learning from other domains~\cite{sun2018transferring}.

In this paper, we investigate a novel strategy that merges the fields of generative models and object detection. While generative models have been proven successful in generating data for training classifiers~\cite{besnier2020dataset}, their application in the realm of object detection has been less explored. The challenge arises when attempting to generate images suitable for object detectors, as it necessitates the representation of targets within specific layouts, adhering to bounding box positions crucial for supervised training. This issue is formulated as \emph{\ac{l2i}} generation. Traditional generative models, however, faced limitations in generating images with such constraints.
To avoid this challenge, some prior approaches employed \acp{gan} for performing \emph{style transfer}, a form of domain adaptation, on the training set~\cite{zhao2022adversarial,huang2018auggan}. Nonetheless, this method relies on an existing training set, whether synthetic or real, and does not truly generate entirely new images.
The first attempt to address the \ac{l2i} problem was made by Layout2Im~\cite{zhao2019image}, but it is only in recent times, notably with GLIGEN~\cite{li2023gligen}, that the generation of constrained images has reached a satisfactory level of quality.

It is worth emphasizing that the existing literature has mostly explored the problem within the realm of generative models, rather than its application to the object detection task. 
The work~\cite{chen2023integrating} also explored the potential of training an object detector using generated images. The authors fine-tuned their generative model on a big real dataset from a very specific domain. 
However, this requires data with time-consuming bounding-box labeling, potentially leading to a cyclical issue where large, labor-intensive data may still be needed to train the generative model instead of the detector.
By contrast, we aim to mitigate this problem, by tackling the data scarcity on generic domains with the aid of pre-trained generative models with \ac{l2i} capabilities, to generate abundant labeled data for free, ready for the training of an object detector. 
Indeed, we intentionally choose to use a pretrained generative model to further alleviate the dependence on real labeled data. Thus, an important novelty of our work is the absence of the generative model training on the specific task domain.

\begin{figure}[t!]
    \centering    
    \includegraphics[width=\textwidth]{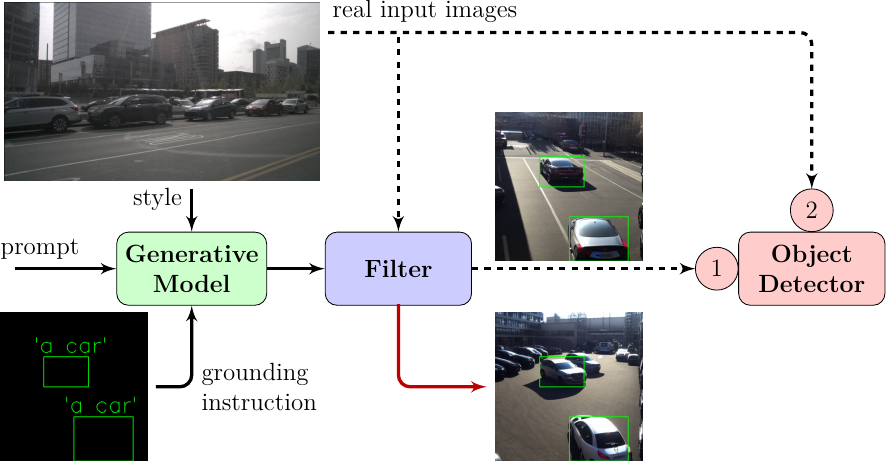}
    \caption{
    \textbf{Transfer learning for object detection with generative models.} We employ a \ac{l2i} pretrained model to generate images for transfer learning to an object detector.
    We can filter out suboptimal generated images based on benchmark metrics. For instance, the image along the red arrow is discarded because the generative model has depicted many cars outside the bounding boxes designated in the grounding instruction.
    With the remaining generated images, we pretrain the object detector, followed by a fine-tuning on the real dataset.
    Dashed lines indicate the data used for training the models.
    }
    \label{fig:pipeline}
\end{figure}

In particular, we exploit GLIGEN, which is peculiarly built upon existing models pretrained
on large-scale web data. This endows GLIGEN with the inherent knowledge derived from other extensive and diverse datasets, facilitating the representation of novel concepts not encountered during its training, aligning seamlessly with our objectives. This sets GLIGEN apart from prior \ac{l2i} generative models, which were typically confined to be trained on datasets tailored to specific tasks. 
Furthermore, the inherent value of this strategy lies in its potential applicability to diverse domains without the explicit requirement of training the generator on each individual domain.

In detail, our architecture, illustrated in \cref{fig:pipeline}, adopts a two-step process. Initially, the object detector undergoes training exclusively on generated images, followed by a subsequent fine-tuning stage using only the limited real dataset. The rationale behind this process is to facilitate transfer learning, enabling the model to exploit a large amount of available data in a source domain, here the domain of generated images, to correctly detect data from the target domain, here the domain of real images, for which the amount of training data is limited. Moreover, we designed an optional filtering strategy to assess the impact of the generated dataset on the object detector and to select the most suitable images for its pretraining.
To assess the effectiveness of our approach, we evaluate it on two different detection domains: cars and fishes in urban and underwater environments, respectively. The reason for this choice is to compare the effectiveness of the proposed approach on a common domain with high data availability (the former) with another domain with data scarcity and interesting for the scientific community (the latter). Nevertheless, the domain choice comes without loss of generality, since our framework can in principle be applied to other common or uncommon domains.

The structure of the paper is as follows: our transfer learning approach with pretrained generative models is detailed in \cref{sec:generationModel,sec:obj-det}. In \cref{sec:filter} the filtering strategies are introduced. \cref{sec:datasets} shows the real datasets selected for our tests, while \cref{sec:results} presents some numerical results. Finally, \cref{sec:conclusion} concludes the paper with some final remarks and outlooks.

\section{Generative model}
\label{sec:generationModel}
Our pipeline is designed to be sufficiently generic, provided that the generator is of the \ac{l2i} type, and it is capable to generalize to diverse domains. 
At the best of our knowledge, only GLIGEN satisfies these requirements in the current state-of-the-art. This is because GLIGEN leverages CLIP, a model trained on vast web-scale data to predict associations between images and corresponding texts, and vice versa. Renowned for its ability to make predictions on concepts that are not explicitly encountered during training, CLIP encoding is leveraged by GLIGEN, achieving the generalized representation abilities we are seeking.

Specifically, the generation process of GLIGEN is guided by an instruction $\textbf{y}=(\textbf{c},\textbf{e})$ that the model follows to create a scene with the desired concepts in specific locations. The instruction comprises a \emph{prompt} \textbf{c}, a written description of the image content, and a \emph{grounding instruction} $\textbf{e}=\left[(e_0,\textbf{l}_0),...,(e_N,\textbf{l}_N)\right]$. Each $\textbf{l}_i$ represents the bounding box coordinates around the object $e_i$ that should be depicted within. These objects, referred to as \emph{grounding entities}, can be described either by text or by a reference real image. 
In \cref{sec:results}, we will present an ablation study comparing the description of grounding entities through text versus real images. This comparison aims to assess which grounding entity type produces targets best suited for detection in the domain of interest.
Regardless of their nature, both kinds of grounding entities are encoded by the pretrained CLIP~\cite{radford2021learning}.
Additionally, associating a reference image as $e_i$ with an image edge as the corresponding $\textbf{l}_i$ facilitates a consistent style transfer from the real to the generated domain. We leverage this feature to achieve backgrounds that closely resemble the originals.

Throughout the remainder of our paper, the grounding instruction corresponds to what in the realm of object detection is referred to as \emph{ground truth}, which is the list of all dataset labels (i.e. bounding boxes and classes of every target).
To automatically choose each $\textbf{l}_i$, we start by collecting the real dataset's ground truth information from a small set of labeled sample shots. Subsequently, we conduct a statistical analysis to estimate the means and variances of the bounding box parameters (i.e. positions, shapes, and number per image) within the original dataset. Following this, we randomly sample from Gaussian distributions using the estimated means and variances, to get plausible parameters for the ground truth to generate.

Therefore, as represented on the left part of \cref{fig:pipeline}, to generate a new image we supply GLIGEN with the following inputs: a text prompt, a desired grounding instruction, an image from the real dataset to get style transfer.
The details on the use of the generation model on example datasets is described in \cref{sec:datasets}.

\section{Object detector}\label{sec:obj-det} 
Our pipeline is general enough to also accept any kind of object detector. For our tests, we opted for the traditional Faster R-CNN~\cite{ren2015faster} which is a region-based model (which gives the letter ``\emph{R}'' in the name). We recall that a Faster R-CNN is mainly composed of two parts, called \emph{backbone} and \ac{rpn}. 

The backbone uses \acp{cnn}, models capable of extracting relevant features, to transform an input image into two-dimensional tensors known as \emph{feature maps}.
The \ac{rpn} overlays the feature maps with a set of bounding boxes, referred to as \emph{anchors}. On each anchor, the network conducts \emph{objectness classification}, which distinguishes generic objects from the background. Subsequently, the anchors containing objects are submitted to a standard neural network classifier for multiclass classification.

In practice, Faster R-CNN is often trained in a transfer learning setting. One common approach involves pretraining the entire model on a standard detection dataset (e.g., COCO~\cite{lin2014microsoft}) and subsequently fine-tuning it on the specific domain of interest.
However, to avoid biases, we choose not to pretrain on other detection datasets, in order to focus on the contribution of generated data only.
Besides, in~\cite{hinterstoisser2018pre} it is demonstrated that training the entire Faster R-CNN on synthetic images leads to significantly poorer performance compared to training on real data. Conversely, freezing a pretrained feature extractor and exclusively training the remaining parts of the detector results in a considerable performance improvement. Therefore, we freeze a pretrained backbone, and we randomly initialize the \ac{rpn} weights.
Finally, it is worth noting that the backbone can be pretrained on a dataset not necessarily tailored for object detection, thus reducing the need for costly labeling. In detail, in our analyses, we opted to use as backbone a ResNet-50~\cite{he2016deep} ---a renowned 50 layers deep \ac{cnn}---pretrained on ImageNet~\cite{russakovsky2015imagenet}, a famous image classification dataset which has become the de facto standard for the pretraining of a \ac{cnn}~\cite{huh2016makes}.
An in-depth description of the ResNet-50 backbone and the Faster R-CNN model are detailed in Appendix A.

Overall, our Faster R-CNN has $41\,370\,911$ weights, of which $23\,454\,912$ belong to the backbone and are frozen. The remaining weights are trained with \ac{sgd} with momentum, which updates the weights $\theta_t$ at time $t$, aiming to minimize a loss function $L$ with the iterative equations \cite{sutskever2013importance}: 
\begin{eqnarray}
     % \left\{
     b_0&=0\ ,\nonumber \\
    b_t&=\mu b_{t-1}+\nabla_\theta L(\theta_{t-1})+\lambda\theta_{t-1}\ ,\\
    \theta_t&=\theta_{t-1}-\gamma b_t\ ,\nonumber 
    % \right.
\end{eqnarray}
involving the so-called \emph{momentum} $\mu$, \emph{weight decay} $\lambda$, \emph{learning rate} $\gamma$, and $\nabla_\theta L$ gradient of $L$. The loss to minimize is formed by the contribution of different terms regarding the positioning of the predicted boxes, the objectness classification and the multiclass classification of each box. In our experiments, we set the value of $\mu=0.9$, while the weight decay $\lambda$ is $0.001$ during the model pretraining on generated data, and $0.01$ for the fine-tuning on the real data. The reason for the latter is that when the model is trained on a limited dataset size, it is more prone to overfitting. We use higher values of $\lambda$, that provide higher \emph{regularization}, i.e. a penalization to reduce the model complexity.

Regarding the learning rate, we adopted two different scheduling strategies, depending on the availability of images for a \emph{validation} set---a portion of the dataset used to monitor the loss value after each \ac{sgd} phase (\emph{epoch}) over the entire training set.
In the fine-tuning stage, due to limited real data, we opted for the common \emph{1x scheduler}. This scheduler does not require a validation set, allowing us to maximize training data usage. 
Conversely, during the pre-training stage, where we can generate as much fake data as needed, we dedicate $15\%$ of the generated data as a validation set.
At the beginning of both stages, the learning rate $\gamma$ is initialized with a starting value of $0.001$. Then, the pretraining step scheduler reduces $\gamma$ by a factor of $10$ if the value of the validation loss does not improve for 5 epochs.
Moreover, we adopt also an \emph{early stopping} strategy where the detector is trained for a maximum of 200 epochs until the validation loss stops decreasing for at least 10 epochs. In the end, we keep the weights of the model at the epoch with the minor validation loss value.
During the fine-tuning step, the \emph{1x} scheduler has a total fixed number of 12 epochs, and $\gamma$ is reduced by a factor of $10$ at epoch 7 and again at epoch 10. The fixed \emph{1x} scheduler should not be considered as a limitation, instead, it has been chosen to achieve a fair comparison, focusing only on the impact of the pretraining stage in the pipeline.

\section{Metrics and filters}\label{sec:filter}
We can evaluate the usefulness of the generation process, described in \cref{sec:generationModel}, through a filter. Its role is to select the best images for the pretraining of the object detector, based on benchmark performance metrics. 
In the current state of the art, the most used metric for evaluating generative models is the \ac{fid}~\cite{heusel2017gans}. It quantifies the similarity and diversity of generated images compared to real data. On the other hand, in the realm of object detection, the main metric to assess the quality of the predictions is known as \ac{map}, which is based on the key concepts of precision and recall. In the following, we define how we calculate \ac{fid}, precision, recall, and the \ac{map}, and we discuss how we use these metrics to filter out suboptimal generated images.

\subsection{\ac{fid} score} \label{sec:fid}
The \ac{fid} score quantifies the similarity between generated and real images, measuring the difference between their feature distributions: $X$ extracted from real images and $Y$ from generated images. These distributions are respectively approximated by two multivariate Gaussian distributions with means $\mu_X$ and $\mu_Y$, and covariances $\Sigma_X$ and $\Sigma_Y$. The \ac{fid} is thus computed\footnote{See the repository \url{https://github.com/mseitzer/pytorch-fid} .} by:
\begin{equation}
    d^2 = ||\mu_X - \mu_Y||^2 + Tr(\Sigma_X + \Sigma_Y - 2\,\sqrt{\Sigma_X\,\Sigma_Y}).
\end{equation}
In detail, we follow the method introduced by \cite{heusel2017gans} and widely adopted in the later literature, which involves extracting features with an Inception-V3 model~\cite{szegedy2016rethinking} pretrained on ImageNet.

Even if \ac{fid} is usually considered as the conventional metric of generative models, its effectiveness in evaluating images used for training object detectors has not been thoroughly explored. Although lower \ac{fid} values indicate better image quality, it remains unknown whether this directly correlates with improved performances for pretraining an object detector. 
To perform this analysis, we introduce a \ac{fid}-based filter, which retains only those images that yield a \ac{fid} score below a certain threshold.
To implement it, we compute the \ac{fid} score between the generated and the real datasets. Then, we iteratively remove one image from the generated dataset, and we re-calculate the \ac{fid} with the real dataset. If this new \ac{fid} is lower than the previous \ac{fid} value, then we definitively remove that image, responsible for distancing the feature distributions between generated and real datasets. Otherwise, the image is kept into the generated dataset.

\subsection{Precision-Recall}\label{sec:precRec}
\begin{figure}[t!]
    \centering
    \begin{subfigure}[b]{0.39\textwidth}
        \includegraphics[width=0.9\textwidth]{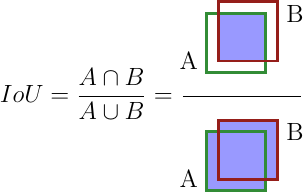}
        \caption{}
        \label{subfig:iouDef}
    \end{subfigure}
    \hfill
    \begin{subfigure}[b]{0.59\textwidth}
        \includegraphics[width=\textwidth]{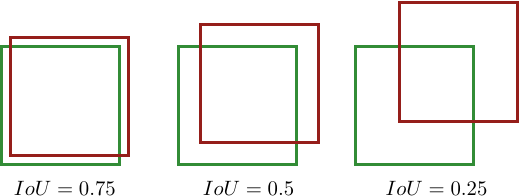}
        \caption{}
        \label{subfig:iouExample}
    \end{subfigure}
    \begin{subfigure}[b]{0.32\textwidth}
         \centering
         \includegraphics[width=\textwidth]{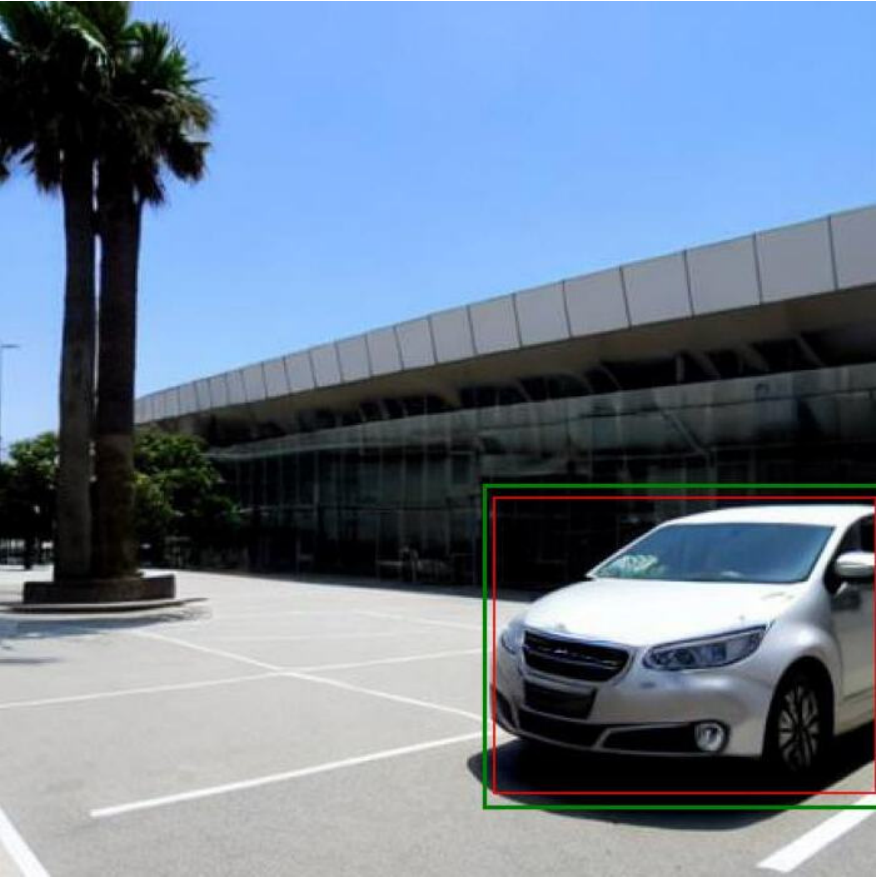}
         \caption{A target is properly generated, and the filter correctly identifies it with $\ac{iou} > 0.5$ as TP, thus Precision and Recall are both 1.}
    \end{subfigure}
    \hfill
    \begin{subfigure}[b]{0.32\textwidth}
         \centering
         \includegraphics[width=\textwidth]{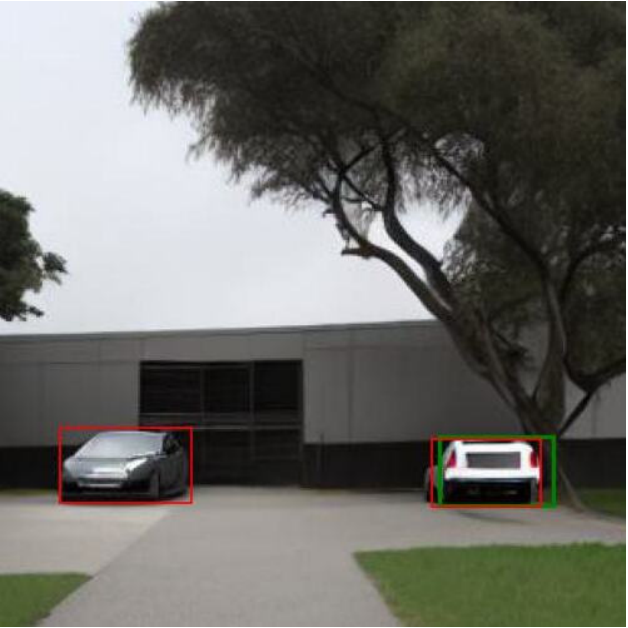}
         \caption{A target is wrongly generated on the left. The filter classifies it with $\ac{iou}=0$ as FP, thus Precision and Recall are 0.5 and 1, respectively.}
         \label{subfig:FP}
    \end{subfigure}
    \hfill
    \begin{subfigure}[b]{0.32\textwidth}
         \centering
         \includegraphics[width=\textwidth]{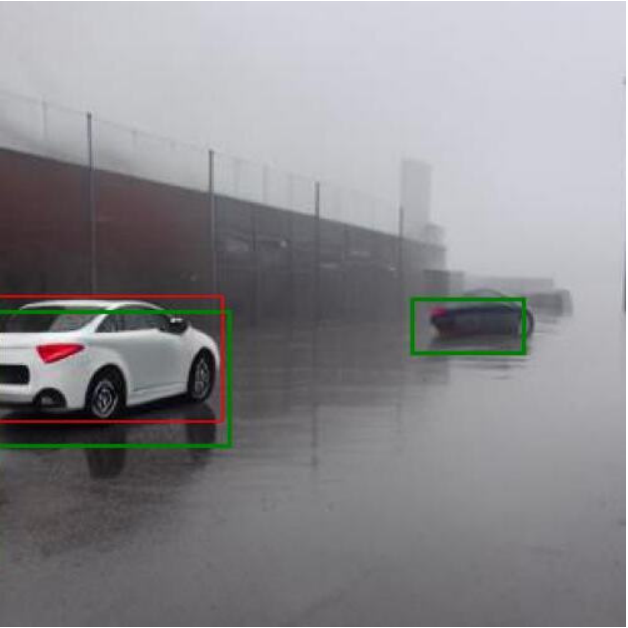}
         \caption{The  target on the right is not well generated. The filter cannot detect it: it counts as FN, thus Precision and Recall are 1 and 0.5, respectively.}
         \label{subfig:FN}
    \end{subfigure}
    \hfill
    \caption{We calculate \ac{iou} as defined in (a) to evaluate the overlap (b) between the ground truth (in \textcolor{colTextGT}{green}) and the predicted boxes (in \textcolor{colTextP}{red}). This is used to implement a Precision-Recall filter to automatically identify faithful representations (c) or discrepancies (d, e) between the intended ground truth for generation and the actual generated images.}
    \label{fig:predictions} 
\end{figure}

To define precision and recall in the context of object detection, we leverage the notion of \emph{\ac{iou}}, which quantifies the similarity between a ground truth and a predicted bounding box. It is computed, as depicted in \cref{subfig:iouDef}, with the ratio between the area of the two boxes intersection and the area of their union.
Its value ranges from 0 to 1: a higher value of \ac{iou} denotes a larger overlap between the boxes as visible in the three examples of \cref{subfig:iouExample}.
By setting an \ac{iou} threshold (typically at 0.5), a prediction box is classified as \emph{True Positive} (TP) if its \ac{iou} score is above the considered threshold, \emph{False Positive} (FP) otherwise. Any failure to detect a ground truth counts as a \emph{False Negative} (FN). 
These values collectively define precision (${prec}$) and recall ($rec$) as
\begin{equation}
    {prec} = \frac{TP}{TP+FP}\ ,\qquad 
    {rec} = \frac{TP}{TP+FN}\ .
\end{equation}
The former measures the correct predictions across all detections and the latter the amount of true targets found among all ground truths.

We propose to employ precision and recall to assess the impact that the generated images may have on the object detector pretraining.
Indeed, the \ac{l2i} models---even the most recent ones---still have some limitations. In particular, they may struggle to adhere to the intended geometric layouts, either drawing targets outside the designated bounding boxes or, conversely, failing to depict the desired objects within them. We want to analyze the generated dataset, identifying potential inconsistencies with the ground truth, in case they may compromise the pretraining of the object detector.
We implement a Precision-Recall based filter with another Faster R-CNN initialized with COCO weights, 
possibly fine-tuned on the limited real dataset. 
This filtering Faster R-CNN is then tested on the generated dataset, computing precision and recall for each image.
By setting thresholds for these metrics, we selectively retain images in which all targets are strictly generated within the specified bounding boxes. 
For instance, setting a prediction threshold at 1, ensures that no target is erroneously generated outside the ground truth constraints, discarding the example in \cref{subfig:FP}. Similarly, a recall threshold at 1 preserves only those images where targets are correctly generated within the specified bounding boxes, discarding the example in \cref{subfig:FN}.

\subsection{Mean Average Precision and F-score}\label{sec:map}
As we mentioned before, the \ac{map} is the principal metric to measure accuracy in the context of object detection. In our analyses, we calculate the \ac{map} on the test set of our real datasets. 

First, for a dataset of $N$ classes, we compute $N$ precision-recall curves, one for each individual class. For a class $i$, this involves ordering the $M_i$ predicted boxes in descending order based on a \emph{confidence} score, which is the probability that the target within $M_i$ is classified as belonging to class $i$.
Then, we compute precision and recall over all images, but considering only the boxes having confidence score greater than a threshold. Such threshold is iteratively decreased, and new values of precision and recall are calculated. Plotting the computed recall values on a x-axis and precision on a y-axis, we get the precision-recall curve for class $i$. 
The \emph{average precision} $AP_i$ is computed as follows~\cite{padillaCITE2020}: 
\begin{align}
    AP_i &= \sum_{j=1}^{M_i-1}(rec_{i,j+1}-rec_{i,j})prec_{int}(rec_{i,j+1}) \\
    prec_{int}(rec_{i,j+1}) &= \max \{prec(rec_{i,k}):rec_{i,k}\ge rec_{i,j+1}\}\nonumber 
\end{align}
where $j=1,\dots,M_i$ are the ordered predicted boxes for class $i$, $rec_{i,j}$ denotes the recall value calculated for class $i$ on the $j$ predicted box, and $prec(rec_{i,j})$ is the precision value corresponding to $rec_{i,j}$.
In other words, the $AP_i$ can be seen as the area under an interpolation of the $i$-th precision-recall curve. At each point, the interpolation considers the maximum precision value for which the recall is greater than or equal to the recall at that specific point on the curve.
Finally, the \ac{map} is the mean of all $AP_i$ over the number of classes $N$. By definition, there is a trade-off between precision and recall: increasing recall will inevitably cause non-relevant results to be detected, decreasing precision. The \ac{map} ranges between 0 and 1: values close to 1 mean that the precision is high also when the recall is increasing. In this case, the number of both false positives and false negatives is low.

Secondly, the performance of our models are evaluated by an F1 score, which considers all the detections having confidence greater than 0.5, and represents both precision and recall in one metric:
\begin{align}
    F_1 &= 2\,\frac{prec\cdot rec}{prec + rec}
\end{align}

\section{Real and generated datasets}\label{sec:datasets}
In this section, we describe the real datasets selected for our tests: NuImages and OzFish. NuImages is representative of a common urban scenario, and serves as a simple testing ground, allowing for initial analyses and illustrative insights into the method's efficacy. OzFish is one of the few publicly available datasets in marine environment. The presence of fish instances, particularly complex to label and detect since they are prone to overlapping, makes OzFish a valuable benchmark for its unique challenges.
Moreover, we also describe the details of how we generate the data used for the pretraining of the detector in the two aforementioned scenarios.

Several datasets have been published to train object detectors in urban environments. For instance, KITTI~\cite{geiger2013vision} is the most widely used for
autonomous driving research. However, the diversity of its recording conditions is relatively low. Instead, NuImages~\cite{caesar2020nuscenes} captures scenes in multiple locations, providing a broader and more varied perspective. Besides, it collects data in both daytime and nighttime scenarios, taking into account various lighting and weather conditions, offering a more realistic and challenging environment for object detection models~\cite{feng2020deep}. 
The dataset is composed by around $75\,000$ images, already divided in $60\, 000$ training samples and $15\,000$ validation samples. It covers up to 25 classes, primarily falling within the broader vehicle and human macro categories. Without loss of generality, we focused our analysis on the car class, aiming for a less complex scenario, in order to give a simpler overview of our proposed method. Consequently, we narrow down our dataset to include only those images containing instances of the car class. Besides, to simulate a situation with limited data available, we make the assumption of using a subset of the large training dataset. Instead, for the test set, we use all the images originally intended as validation and that contains at least a car instance, to achieve a more statistically meaningful result. In summary, our selected dataset contains 4\,500 training images at most, and $9\,073$ test images.

To generate each image for the pretraining dataset in the urban scenario, we recall that we need a prompt \textbf{c} and a grounding instruction $\textbf{e}=\left[(e_0,\textbf{l}_0),...,(e_N,\textbf{l}_N)\right]$, whose first value $e_0$ is specifically the real image used for style transfer.
With $N$ cars to generate, the other grounding entities $e_i$, for $i = 1,\dots,N$, are described by the phrase "a car", as shown in \cref{fig:pipeline}. In particular, we use 3\,644 training images for the style transfer, which may seem a large number, but it actually requires no human labeling intervention.
Determining the most effective prompt for guiding the generation process falls within the field of study known as \textit{prompt engineering}, but this is beyond the scope of our current investigation. Instead, to evaluate the potentialities from a non-expert perspective, we opt for the basic prompt "$N$ cars in a urban environment, highly photorealistic", where $N$ is the desired number of cars in a single image. 
It is crucial to note that although we have demonstrated a generation focused on a single class example, our methodology can be effortlessly extended to a multi-class scenario by appropriately setting different class phrases for the $e_i$ in the grounding instruction.

For our second test scenario, in marine environment, we remark that the task of acquiring high-quality labeled underwater images is challenged by two main factors~\cite{yang2021computer}. On one hand, there are several environment uncontrolled variables, such as water turbidity, uneven illumination, and low contrast. On the other, also intrinsic fish characteristics contributes to the task difficulty, including intra-class variations, color similarity with the background, occlusions, and arbitrary unusual swimming positions.
Besides, the rich biodiversity adds complexity to creating a universal dataset for marine ecosystem observation, limiting model applicability across different underwater areas. 
This is in contrast with the \ac{ml} requirement for great quantity of good quality training data, and existing datasets, as summarized by~\cite{ditria2021annotated}, often fall short in addressing this requirement. 
For instance, Fish4Knowledge~\cite{fisher2016fish4knowledge} and Rockfish~\cite{cutter2015automated} do not tackle the object detection challenge, mainly addressing it by applying classic post-processing cropping methods, so that each fish results centered and can be identified with a mere object classification approach.
The DeepFish~\cite{saleh2020realistic} dataset surpasses the previous ones by also offering a \textit{semantic segmentation} of fish instances, i.e. the annotation of every pixel belonging to the fish, denoted as \textit{mask}.
A Seagrass dataset was used in~\cite{ditria2020automating} to demonstrate that \ac{ml} approaches result in an increased abundance detection compared to human experts. 
However, all previous dataset presents limitations, either due to poor image resolution (Fish4Knowledge) or because they have a low number of fishes per image (1 at most for DeepFish and RockFish, not more than 18 on Seagrass).

The OzFish~\cite{ozfish} dataset has significantly increased the number of annotated boxes per image, reaching up to hundreds. As stated in~\cite{veiga2022autonomous}, this comes with incorrect or missing bounding box annotations. Examples include overlapped annotations on the same target, negligible small bounding boxes, and mislabeling of objects like reefs and floating ropes annotated as fish. Despite these imperfections, the data remains valuable, since the images are more extensively labeled compared to previous datasets, and they also exhibit greater variability in environments and fish positioning, allowing us to address challenges related to turbidity and overlapping.
In particular, OzFish consists of around $1\,800$ images and containing around $45\,000$ bounding box annotations. 
We partitioned the images using an 85\% to 15\% ratio, resulting in 1\,500 images for the training set and 265 images for the test set.

\begin{figure}[t!]
    \centering
     \begin{subfigure}[b]{0.32\textwidth}
         \centering
         \includegraphics[width=\textwidth]{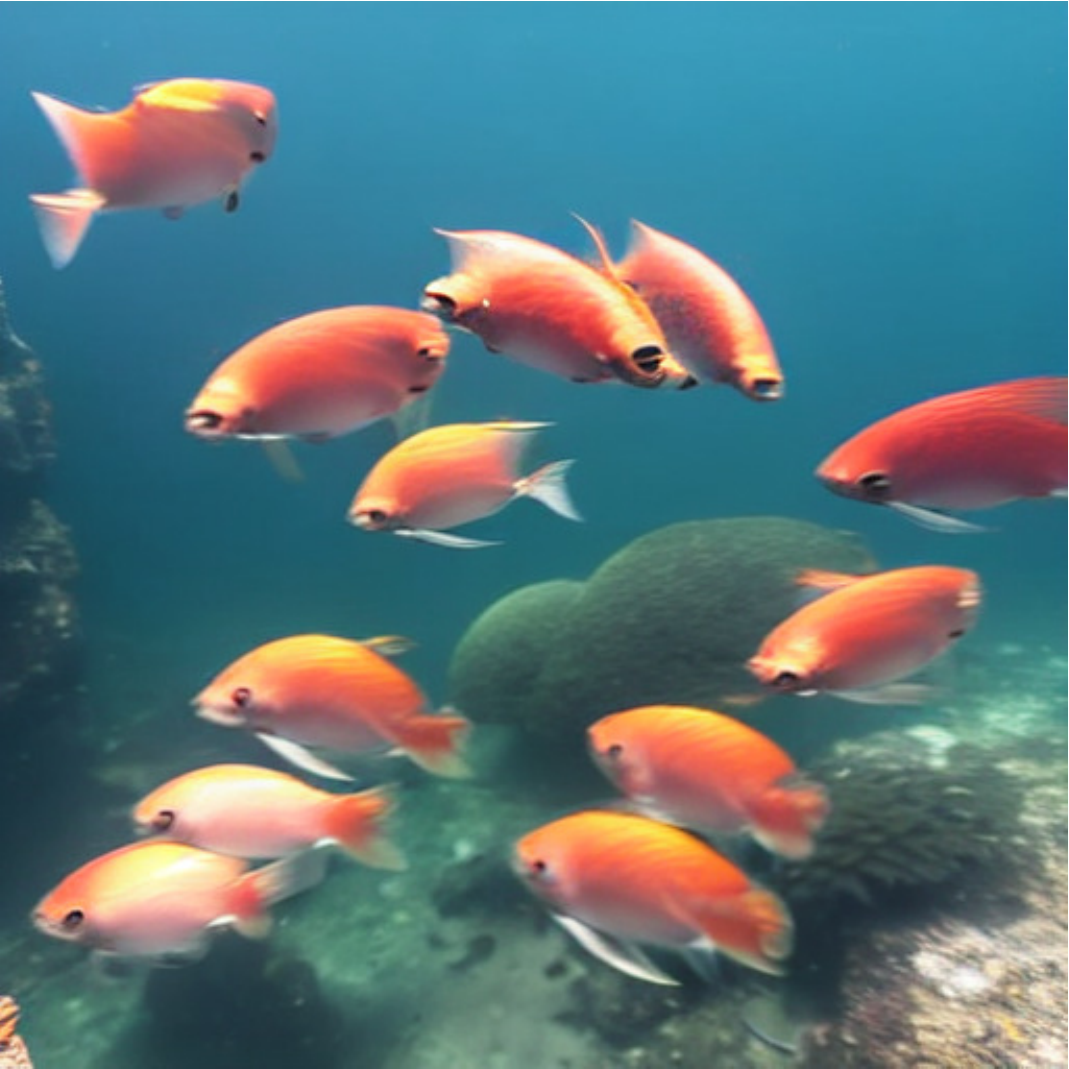}
         \caption{}
         \label{fig:fish_txt}
     \end{subfigure}
     \hfill
     \begin{subfigure}[b]{0.32\textwidth}
         \centering
         \includegraphics[width=\textwidth]{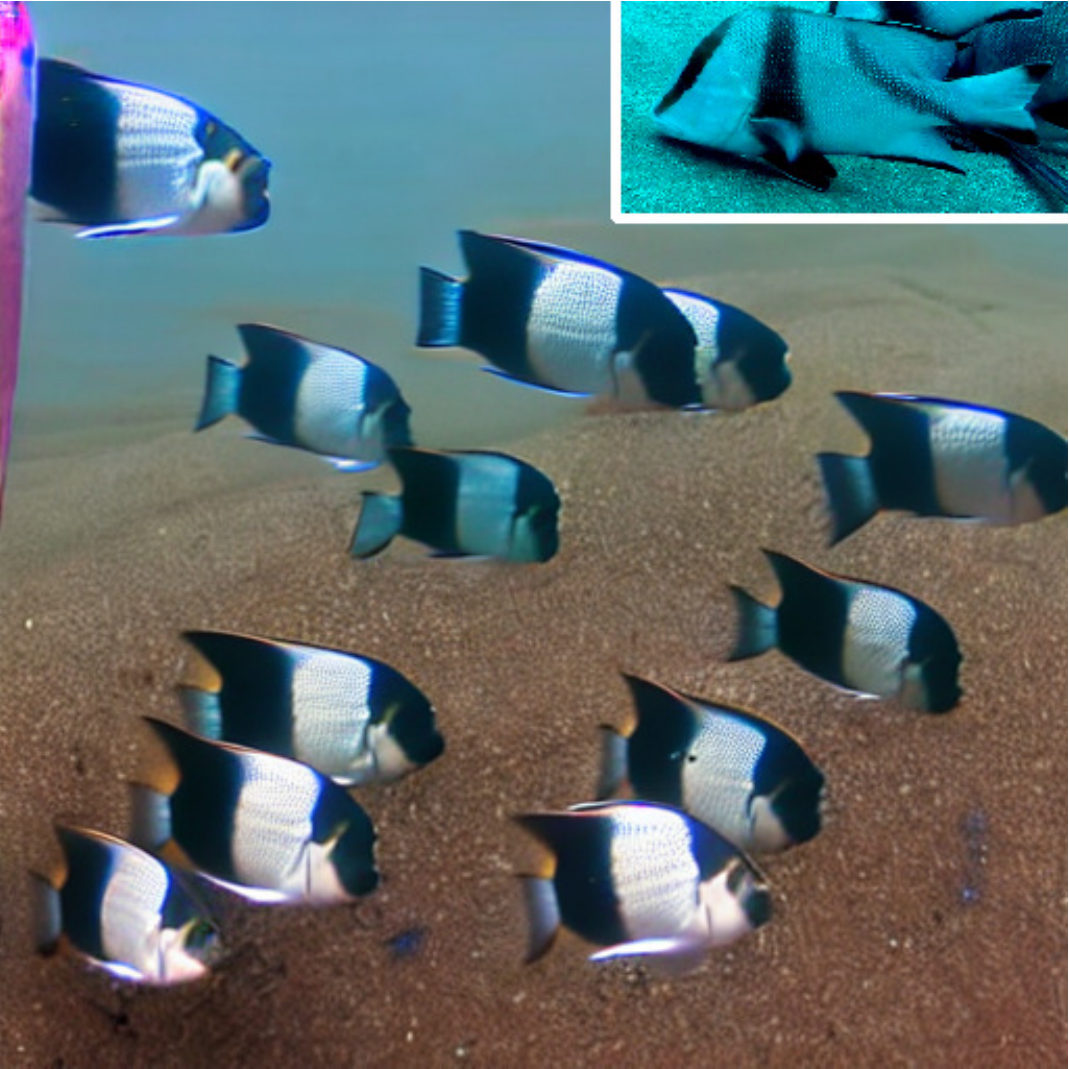}
         \caption{}
         \label{fig:fish_img}
     \end{subfigure}
     \hfill
     \begin{subfigure}[b]{0.32\textwidth}
         \centering
         \includegraphics[width=\textwidth]{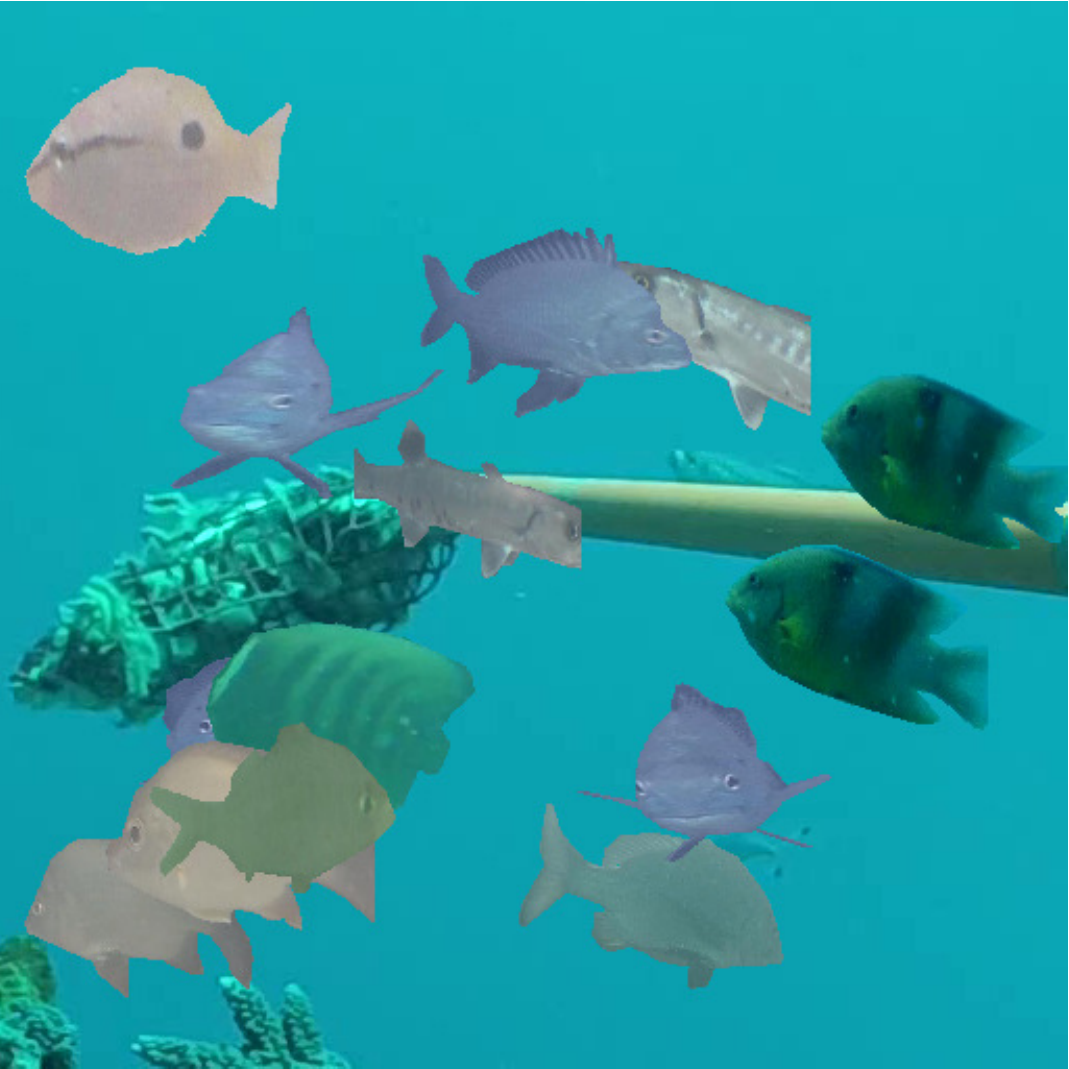}
         \caption{}
         \label{fig:fish_msk}
     \end{subfigure}
     \caption{Fish instances are positioned in identical bounding boxes using three different methods: employing as the GLIGEN grounding entities the text phrase "a fish" (a) or an image of a real fish, as shown in the box in the top right corner of (b); pasting DeepFish masks onto an OzFish background (c).}
     \label{fig:fish_types}
\end{figure}

For the generated dataset, we decided to explore two distinct approaches, comparing a grounding instruction that relies on text descriptions with another one where example images are used. In the former, each target grounding entity is defined simply by the phrase "a fish" (an example is visible in \cref{fig:fish_txt}); in the latter, we employ as grounding entities one image selected from a set of $16$ close-ups of different fish species from the training set (as in \cref{fig:fish_img}). In both cases, one sample from a set of $128$ training pictures without fishes is employed to condition the background style of a single image. The chosen prompt is "blurry fishes in motion, foggy underwater professional photography". 
In \cref{sec:ozfish}, we will compare these two kinds of generated dataset with a common state-of-the-art technique, where a synthetic image is created by copying masks of the real objects and pasting them onto a real background, as in \cref{fig:fish_msk}. To achieve this, we leverage the DeepFish dataset that provides semantic segmentations of fishes. We obtained $207$ masks of around $10$ different species, and we pasted them on the same $128$ images used for the style transfer in the previous approaches. We note that, despite the diversity between OzFish and DeepFish datasets, both feature tropical Australian species.
For these three types of approaches, we defined identical desired bounding box positions for all the fish instances to ensure more comparable training scenarios. However, as discussed in \cref{sec:filter}, the generation model might struggle to generate all the targets, particularly in cases of overlap, which is a challenge not encountered by the copy-paste method (as an example of this issue, compare the fishes in the bottom left corner between the three images in \cref{fig:fish_types}). Finally, we observe that the fish instances in these three datasets exhibit increasing levels of complexity. Creating a fish with the first method merely requires a text phrase. The second approach demands capturing well-centered pictures of the real targets of interest, which is still a relatively simple task. The third one needs semantic masks of the objects, making it the most laborious method, involving more human intervention than the other two.

For all the aforementioned datasets, both real and generated, we exclude the use of small bounding boxes, specifically ignoring those with an area that is less than 0.2\% of the image area. Indeed, it is well-known in the literature that generative models struggle with generating small targets~\cite{bosquet2023full}, a problem that falls into another research field and is beyond the scope of this study.

To generate the datasets, we have used
two computational servers: one with 88 Intel Xeon E5-2699 v4 CPU, 94 Gb of RAM and a Nvidia TITAN RTX GPU with 24 Gb of memory; the second one with 32 AMD Ryzen 9 5950X CPU, 126 Gb of RAM and a Nvidia GeForce RTX 3090 GPU with 24 Gb of memory. The execution time to generate each single image is around 37 seconds on the first machine and 21 seconds on the second one.

\section{Results and discussion}\label{sec:results}
\begin{figure}[t!]
    \centering
     \begin{subfigure}[b]{0.49\textwidth}
         \centering         \includegraphics[width=\textwidth]{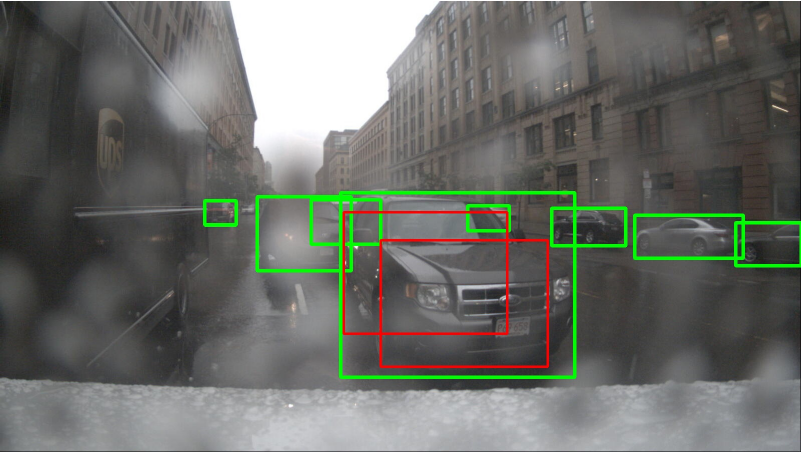}
         \caption{}
         \label{fig:nu_real_only}
     \end{subfigure}
     \hfill
     \begin{subfigure}[b]{0.49\textwidth}
         \centering         \includegraphics[width=\textwidth]{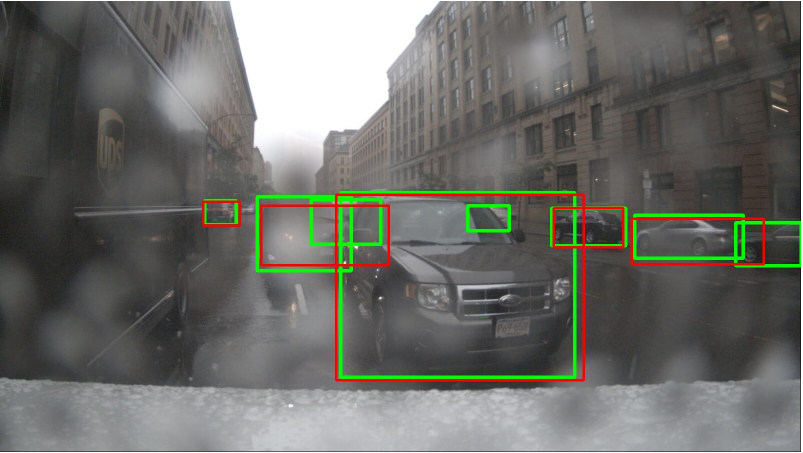}
         \caption{}
         \label{fig:nu_pretrained}
     \end{subfigure}

     \begin{subfigure}[b]{0.49\textwidth}
         \centering         \includegraphics[width=\textwidth]{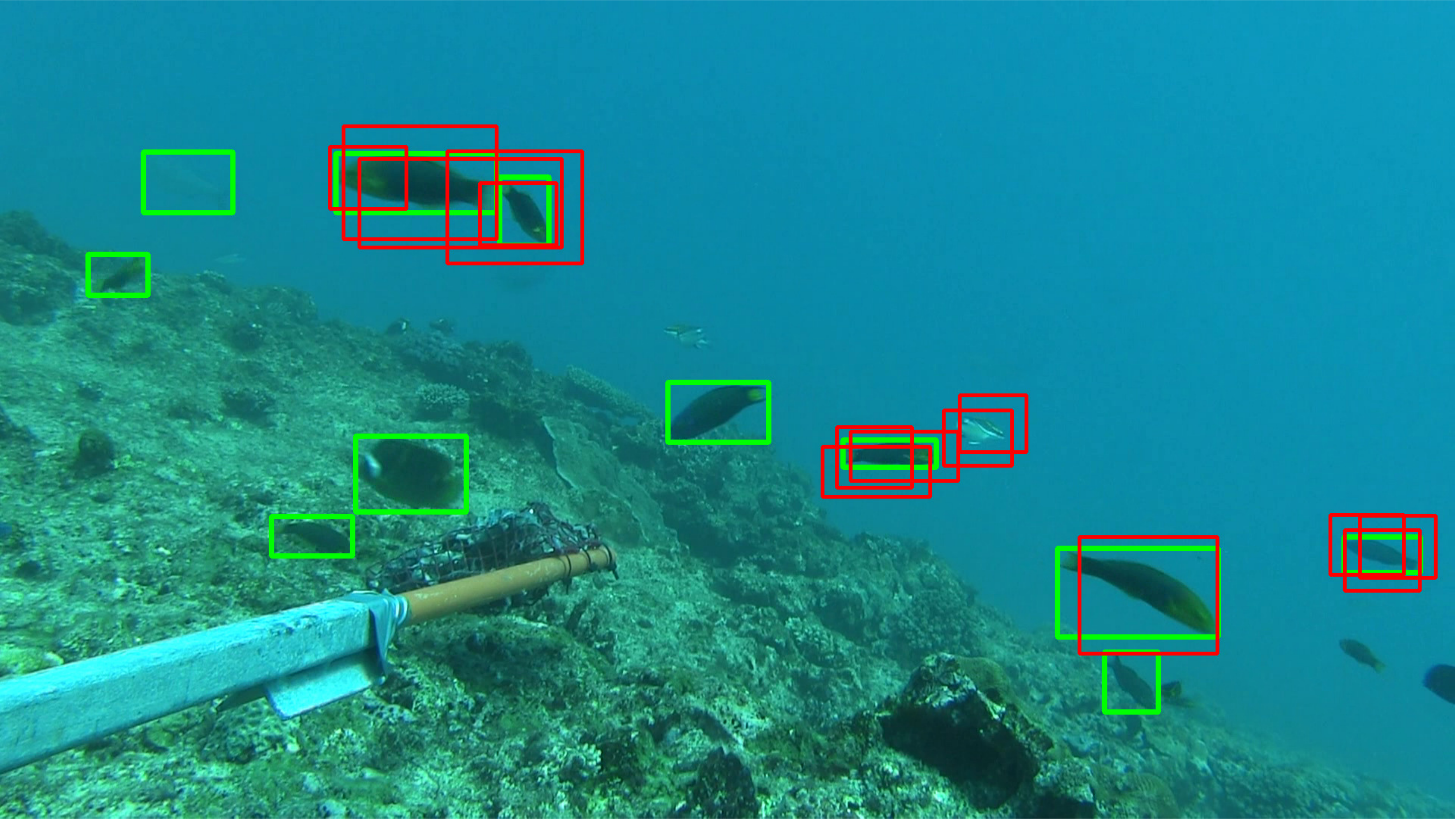}
         \caption{}
         \label{fig:oz_real_only}
     \end{subfigure}
     \hfill
     \begin{subfigure}[b]{0.49\textwidth}
         \centering         \includegraphics[width=\textwidth]{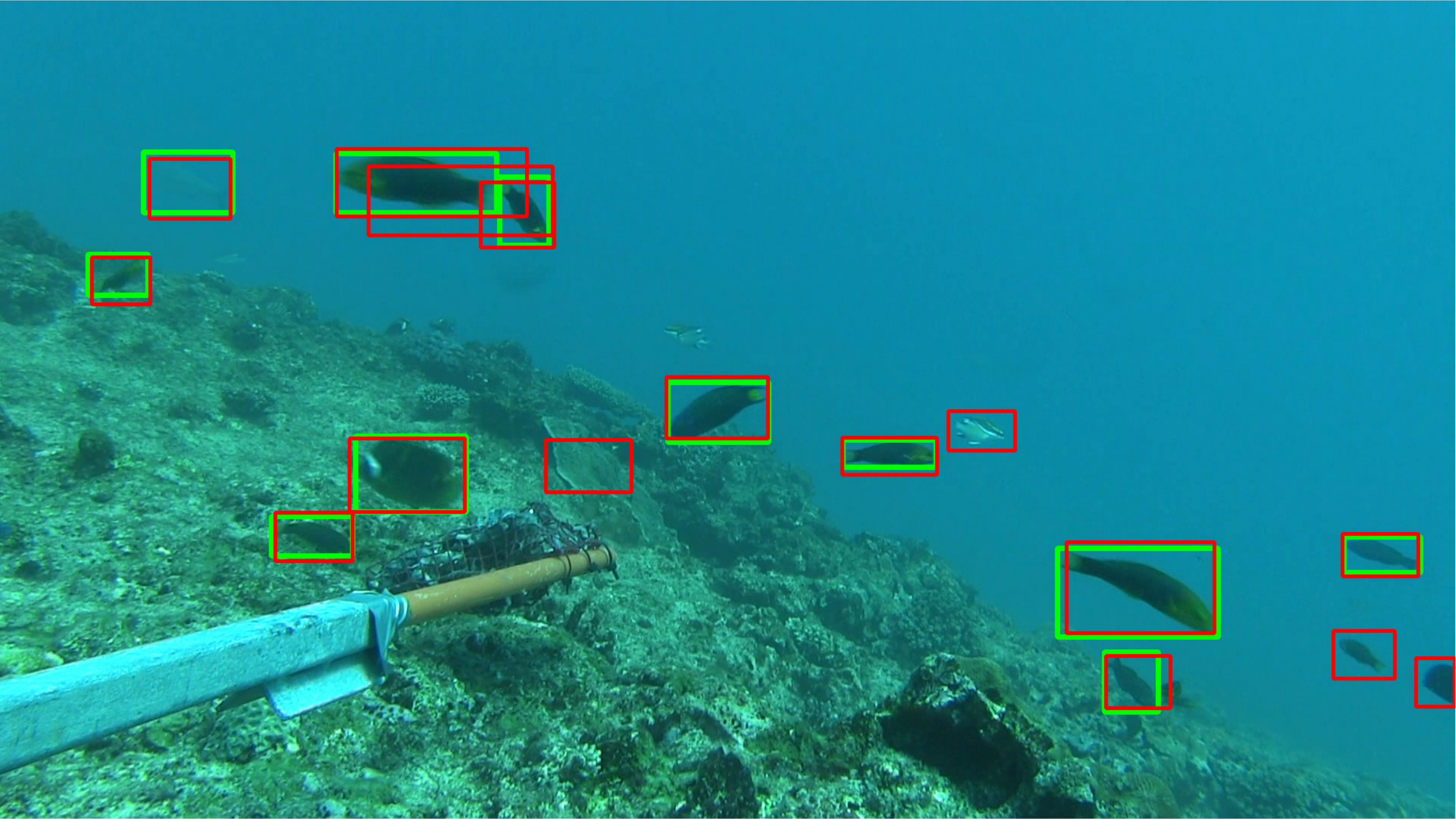}
         \caption{}
         \label{fig:oz_pretrained}
     \end{subfigure}
     
     \caption{Side-by-side illustrations comparing examples of object detector predictions (in \textcolor{colTextP}{red}) with the corresponding testing ground truth (in \textcolor{colTextGT}{green}) under our two scenarios. In the left column (a, c), Faster R-CNN models are initialized with ImageNet and fine-tuned using a minimal dataset of 300 real training images. On the right, models undergo a preliminary pretraining on 9\,000 generated images (from the image grounding dataset in the OzFish case), followed by a fine-tuning on the same 300 real images.
     We can observe that our approach shows increased performances in finding more targets and refining redundant detection boxes, even in rainy or turbid environments. 
     }
     \label{fig:predictionsExamples}
\end{figure}

To validate our pipeline, we have implemented it in PyTorch, and we have used the same servers specified in \cref{sec:datasets} to train several models on both NuImages and OzFish datasets. The training times are around 0.7 seconds per iteration on the first machine, and 0.3 seconds per iteration on the second one. In the following, the details of our analyses are explained. In \cref{fig:predictionsExamples} we show some examples of prediction in both scenarios.

\subsection{NuImages---reducing the need for real data}\label{sec:nuimages}
\begin{figure}
    \centering
    \includegraphics[width=0.95\textwidth]{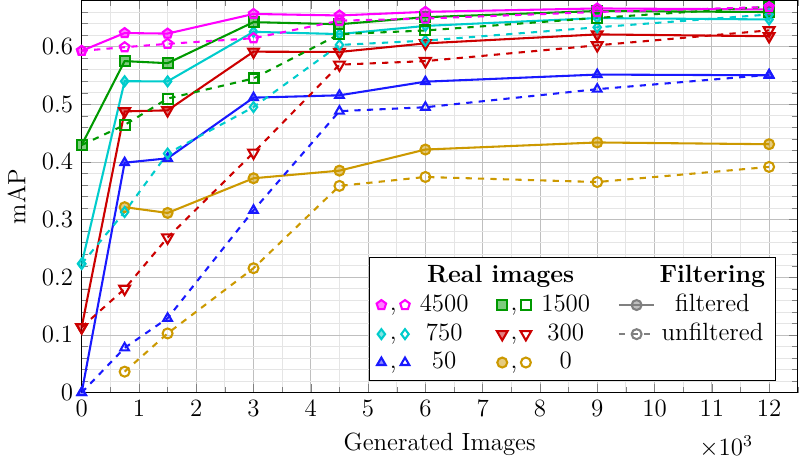}
    \caption{Results of the object detection \ac{map} evaluated on the same NuImages test set, illustrating the impact of pretraining with varying numbers of generated images (x-axis) and subsequent training with different quantities of real images from the NuImages training set (curve color and marker shape). Solid lines with filled markers represent results with pretraining on filtered images, while dashed lines with empty markers depict pretraining on unfiltered images (in the right part of the legend we report only one marker shape for compactness).}
    \label{fig:mapNuimages}
\end{figure}
On NuImages, we show the effectiveness of our transfer learning approach by evaluating the \ac{map} performances of our object detector under different setup conditions, all tested on the same set of 9\,073 real images. We also examine the impact of filters on the model training. The Faster R-CNN is pretrained using varying amounts of our generated images---specifically, 750, 1\,500, 3\,000, 4\,500, 6\,000, 9\,000 and 12\,000. For comparison, we report the case where the generated images are not used during pretraining. During this phase, the previously specified quantities correspond to an 85\% of the total generated images used for each configuration, while the remaining 15\% is kept for validation and scheduling purposes. For instance, the second configuration is trained on 1\,500 images from the generated set and validated on 265 other images.
Subsequently, each of these pretrained models undergoes fine-tuning with various amounts of real images---namely, 50, 300, 750, 1\,500, and 4\,500. We also considered a model only pretrained on generated images, without any fine-tuning on the real dataset. 

In \cref{fig:mapNuimages}, the x-axis represents the number of generated images used for the initial pretraining, while different colors and marker shapes indicates the numbers of real images used for the fine-tuning step. 
The y-axis shows the corresponding \ac{map} performances on the real test set for each configuration. 
Besides, to evaluate the usefulness of our filter, we followed two distinct approaches: one involving the selection of generated images through a Precision-Recall filter (solid curves and filled markers), and the other employing all generated images, regardless of their quality (dashed curves and empty markers). Specifically, for both precision and recall, we set the filtering thresholds to 1, aiming to preserve only the most accurate representations of the ground truth during the generation process (this process discarded around 56\% of all the images).

Several key observations can be drawn. Unsurprisingly, increasing the number of real images during the fine-tuning results in higher \ac{map} values. In fact, the curves where a higher number of real images is used, consistently appear on top of the ones with fewer real images. 
The first observation regarding our approach is that the integration of generated images successfully compensates for the scarcity of real data. For instance, 300 real images combined with 9\,000 generated images exhibit performance equivalent to the full dataset of 4\,500 real images.
More surprisingly, a combination of 300 real images and 750 filtered generated images yields better results than using only 1500 real images. This finding is quite remarkable, suggesting that the contribution of the pretraining is crucial to obtain a good starting point for the fine-tuning of the weights with the real images. This assertion is also supported by the curve indicated with 0 real images, where favorable \ac{map} values can be achieved solely with generated images, even without fine-tuning on real ones. For example, training the detector with 9\,000 filtered generated images is comparable to the performance obtained on 1\,500 real images. We assume that the good impact of the generated images can be attributed to GLIGEN being trained also with knowledge from the urban domain.

The second observation that we can deduce is that the filter effectively condenses the information from a multitude of generated images into few important ones. Interestingly, the filtered images do not allow for better performance over the unfiltered ones in the long run. In fact, the \ac{map} converges to similar values when pretraining with 12\,000 images, both filtered and unfiltered. We make the hypothesis that, for large numbers of generated training images, the model autonomously learns how to ignore the imperfect labeling occurring in the generated images. Conversely, when the size of the training set is small, the model may overfit on these inaccurate data.

In another test, we also evaluate the influence of the \ac{fid} score on the training of the object detector. We arranged the unfiltered generated images into three mutually exclusive subsets with varying \ac{fid} scores: 47, 105, and 148. Each subset consisted of 3\,000 images for pretraining, with an additional 530 images used for validation.  All these configurations are then involved in a fine-tuning step on the same 300 real images of \cref{fig:mapNuimages}.
At the end, we found that there are no significant improvements in training the detector on datasets with different \ac{fid} scores. Specifically, for all three cases we obtained really similar test detection \acp{map} comparable with the results of \cref{fig:mapNuimages}.
This is an evidence that, even if the \ac{fid} score serves as a valuable metric for assessing the overall similarity between real and generated distributions, it may not be the most suitable metric to measure the impact of generated data in object detection tasks. Specifically, it is important to recognize that FID only assesses the image quality independently from specific downstream tasks like object detection. Indeed, the \ac{fid} calculates distribution distances considering the images as a whole, rather than focusing on the quality of individual targets, which is the primary concern of the object detector's feature extractor.
For example, we can hypothetically consider two distinct detection datasets having exactly identical images, but one with the correct labeling and the other one with a completely erroneous one. In this scenario, the \ac{fid} distance between the two datasets would be 0, but their impact on the training of an object detector would be completely different, catastrophic for the wrongly labelled dataset.

\subsection{OzFish---comparing with different state-of-the-art techniques and testing universality}\label{sec:ozfish}

\begin{figure}
    \centering
    \includegraphics[width=0.95\textwidth]{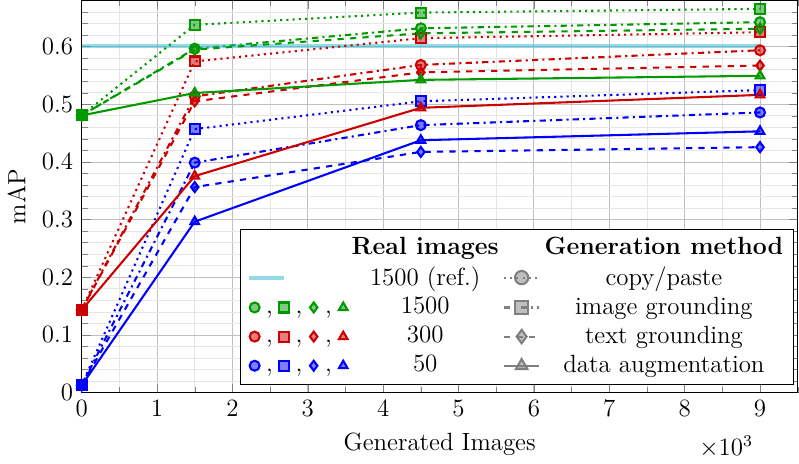}
    \caption{Object detection \ac{map} results on the OzFish real test set, for models pretrained on different quantities of unfiltered generated images (x-axis) and fine-tuned on varying numbers of OzFish training images (specified by the curve color). Dotted curves with round markers indicate models pretrained on synthetic copy-paste images. Dot-dashed and dashed lines, with squares and diamond markers, represent the use of images and text as grounding entities, respectively. The cyan constant solid line at around $0.6$ \ac{map} reports for reference the performance of a model pretrained on COCO and fine-tuned on all the 1\,500 OzFish training images. The colored solid lines with triangular markers are references that use a standard data augmentation approach.
    }
    \label{fig:mapOzfish}
\end{figure}

In this section we discuss the application of our pipeline to the domain of OzFish. We present a comparison between our method and other prominent state-of-the-art approaches. Moreover, in this domain we decided not to focus on the effects of the filter because we observed a fast convergence already using the unfiltered images.
We employed the three datasets outlined in \cref{sec:datasets} and exemplified in \cref{fig:fish_types}. Similar to the generated dataset used for the NuImages scenario, we maintained an 85\% of the generated images for pretraining and reserved the remaining 15\% for validation. As in \cref{fig:mapNuimages}, in \cref{fig:mapOzfish} we plot the \ac{map} trend for an increasing number of pretraining generated images (x-axis) and fine-tuning real images (curve color). For all cases, the \ac{map} is computed on the same real test set of 265 OzFish images.

The first insightful observation arises when comparing various image generation strategies. We observe that each method's curve appears completely above the previous ones. Specifically, the state-of-the-art copy-paste approach (dotted curve with round markers) exhibits the highest performance, followed by the cases where grounding entities are images (dot-dashed curve with square markers) and text (dashed curve with diamond markers). This is probably because the copy-paste dataset incorporates true fishes features, which may prove more effective than the generated ones. 
However, it is crucial to note that the curves appear to be sorted by an increasing level of their generation complexity. It is important to recall that the copy-paste approach synthesizes images by relying on mask segmentation, a process that is inherently resource-intensive, even more than labeling a dataset for object detection. In contrast, our methods stand out for their simplicity, requiring only target pictures for image grounding or even simpler text descriptions. For instance, under the configuration with 9\,000 pretraining images and 300 real images, the \ac{map} gap between text grounding and copy-paste methods is just 0.05, which is quite tolerable, considering the substantial reduction in the burden on human intervention.

To further investigate the contribution of fish features to \ac{map} performance, we calculate the \ac{fid} values between each of these datasets and the real Ozfish dataset, resulting in values of 70, 90, and 100 for the copy-paste, image grounding, and text grounding approaches, respectively. Indeed, the copy-paste method exhibits the lowest \ac{fid} (70), indicating a feature distribution closest to that of OzFish. In this scenario, fine-tuning with 50 OzFish images (blue curve) shows a 0.1 \ac{map} gap between copy-paste and text grounding pretraining, suggesting that datasets with lower \ac{fid} values contribute more effectively to the object detector pretraining.
This finding appears to differ from the \ac{fid} filter results discussed in \cref{sec:nuimages}. In that context, three datasets were created filtering out images generated with the same method, using identical text grounding entities and prompts, resulting in varying \ac{fid} values (47, 105, 148). However, used during detector pretraining, these datasets yielded comparable \ac{map} values, implying that \ac{fid} was not effective in filtering detrimental images.
This apparently contrasting result might find an explanation by considering that the \ac{fid} can be influenced by many factors whose impact on the \ac{map} is not necessarily equal. For instance, image background and targets may have features that do not hold the same weight in the task of pretraining an object detector. 
It is essential to note that our detector has a frozen backbone and is exclusively trained on the \ac{rpn} detection portion, which focuses solely on features within the target's bounding boxes and not on the background. In the NuImages case, where target features are generated with the same instructions, \ac{fid} differences probably arise mainly from variations in background features, irrelevant to the detection task. Therefore, using a \ac{fid}-based filter on those generated datasets may discard features not relevant for the \ac{rpn} training. Conversely, in the case of OzFish, the three datasets inherently contain differently generated targets, resulting in \ac{fid} differences that may also be caused by variations in the targets' features, contributing to both \ac{fid} variation and \ac{rpn} training performances.

We also compare our model with the classical data augmentation techniques illustrated in \cref{sec:introduction} (colored solid lines with triangular markers in \cref{fig:mapOzfish}). We augment independently each of the considered real sub-datasets, and we directly train the object detector (pretrained only on ImageNet) on a mix of real and augmented images, without our 2-step transfer learning.
Moreover, we establish another reference configuration whose weights are initialized based on the COCO dataset as in~\cite{ren2015faster}. This allows to benchmark the performances of our generated datasets against the state-of-the-art method to address data scarcity, that is to perform transfer learning using a model pretrained on the most prominent dataset for object detection.
In detail, the reference model is pretrained on ImageNet and then on COCO.  We additionally fine-tune it on the whole OzFish dataset, keeping also in this case the backbone frozen as detailed in \cref{sec:obj-det}. 
Our model surprisingly approaches the performance of the COCO-pretrained reference (cyan constant solid line at around $0.6$ \ac{map} in \cref{fig:mapOzfish}) with only 300 OzFish examples (red curve), reducing the number of real labeled images by almost three orders of magnitude: from 118K COCO images plus 1\,500 OzFish images to only 300 OzFish images supported by 9\,000 generated ones.
Besides, with this specific training setting, when using the entire dataset of 1\,500 real images, we manage to surpass the reference with only 1\,500 generated images (green curve). These are insights that the generated data is better suited for the transfer learning task in this context, compared to the general-purpose COCO dataset. This is encouraging for the widespread adoption of generative models in object detection tasks.

\begin{figure}[t!]
    \centering
    \includegraphics[width=0.95\textwidth]{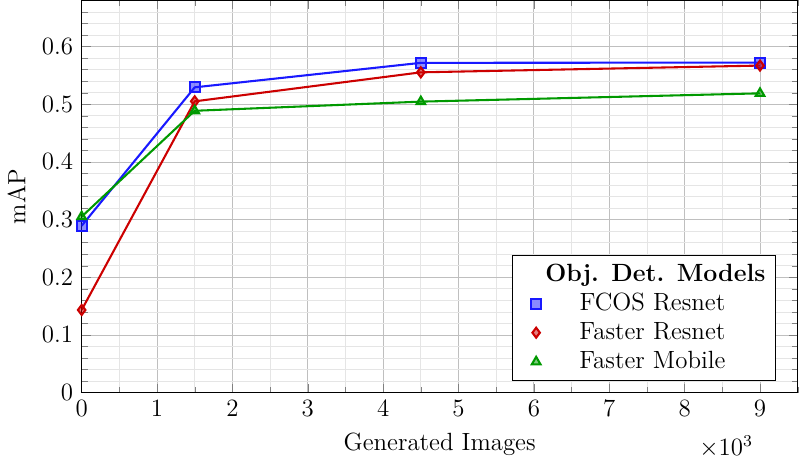}
    \caption{Comparison of the Faster R-CNN model used in \cref{fig:mapOzfish} with other models. All tests use 300 real images with pretraining on text-grounded generated images. }
    \label{fig:modelComparison}
\end{figure}

Furthermore, in \cref{fig:modelComparison}, we test the universality of our framework by comparing our approach across different object detectors. 
We recall that the object detector used in previous evaluations is a Faster R-CNN with a ResNet-50 backbone (labeled as \emph{Faster Resnet} in the legend). This is compared with two different scenarios. In the first, the same Faster R-CNN model is used, but with a different backbone network: MobileNet\_v3~\cite{howard2019searching} (labeled as \emph{Faster Mobile}). The second scenario employs the same ResNet-50 backbone, but with a completely different detection model: \ac{fcos}~\cite{tian1904fcos} (labeled as \emph{FCOS Resnet}). Technical details of these models can be found in the Appendix A. We observe that all the three models exhibit a similar behavior, demonstrating that our approach is independent of the chosen object detector and highlighting the beneficial effect of pretraining on generated images in every scenario.
Finally, all the main metrics (\ac{map}, precision, recall and F1 score) for some relevant trainings of the object detector on the OzFish dataset are reported in \cref{tab:metrics}, along with the results on NuImages.

\begin{table}[h!]
    \centering
    \caption{Metrics for the two considered datasets, for 50 and $1\,500$ real images and for 0 and $9\,000$ generated images. The confidence threshold to compute precision, recall and F1 is fixed at 0.5. The OzFish generated dataset is the image grounding one.}
    \label{tab:metrics}
    \begin{tabular}{|c|c|c||c|c|c|c|}
        \hline
        \textbf{Dataset} & \textbf{Real im.} & \textbf{Generated im.} & \textbf{Prec.} & \textbf{Rec.} & \textbf{F1} & \textbf{mAP} \\
        \hhline{=======}
        \multirow{4}{*}{NuImages} & \multirow{2}{*}{50}       & 0        & 0    & 0     & -   & 0.74e-6\\
        \cline{3-7}
                                  &                           & $9\,000$ & 0.51 & 0.56  & 0.53  & 0.55\\
        \cline{2-7}
                                  & \multirow{2}{*}{$1\,500$} & 0        & 0.35 & 0.54  & 0.42  & 0.43\\
        \cline{3-7}
                                  &                           & $9\,000$ & 0.52 & 0.67  & 0.59  & 0.66\\
        \hline
        \multirow{4}{*}{OzFish}   & \multirow{2}{*}{50}       & 0        & 0    & 0     & -   & 0.024\\
        \cline{3-7}
                                  &                           & $9\,000$ & 0.48 & 0.51  & 0.50  & 0.48\\
        \cline{2-7}
                                  & \multirow{2}{*}{$1\,500$} & 0        & 0.41 & 0.54  & 0.46  & 0.46\\
        \cline{3-7}
                                  &                           & $9\,000$ & 0.57 & 0.65  & 0.61  & 0.64\\
        \hline
    \end{tabular}
\end{table}

\section{Conclusion}
\label{sec:conclusion}
We have proposed a method that leverages GLIGEN, a generative model pretrained on extensive and diverse datasets, to generate images with targets inside bounding boxes for the pretraining of an object detector and subsequent transfer learning to the real domain. We have demonstrated remarkable results on both NuImages and OzFish datasets. We have showed that our method allows to achieve good detection performance with just few hundreds of real images, comparable to training the model on thousands of real images.
Furthermore, our method significantly reduces the human effort involved in the annotation process, limiting the need for labor-intensive labeling tasks.

In our work we address the generalization capability of the proposed approach by comparing a common-domain dataset of cars in urban environment with a more specific and uncommon dataset of fishes in underwater environment. Besides, it is plausible that our framework can be generalizable even to more complex scenarios. Indeed, GLIGEN leverages CLIP, i.e. a model trained on vast and diverse datasets that encompass categories beyond just underwater fish and urban cars. CLIP requires only image-text pairs, which are more easy to obtain compared to labor-intensive annotations like bounding boxes. GLIGEN builds upon CLIP’s strong representation capabilities and learns only the additional ability to depict objects within the desired bounding boxes. This separation of capabilities suggests that the two aspects are relatively independent. Therefore, in more niche scenarios, such as on medical images, it should be sufficient to fine-tune the underlying CLIP model~\cite{zhao2023clip}.

During the preparation of our manuscript, an independent paper has appeared online as preprint in Ref.~\cite{chen2023integrating}, exploring the potential of training an object detector using generated images. However, their generative model was trained on a substantial amount of real data from a specific dataset, thus without addressing the issue of limited datasets from a generic domain that we have considered here. 
Besides, while via their model they primarily report the \ac{fid} as the standard metric to evaluate generative models, we here also explore how \ac{fid} impacts on \ac{map}. While our findings are in their preliminary stages and comprise hypotheses, they open new possibilities for understanding the relationship between \ac{fid} and \ac{map}, hence providing an intriguing direction for further exploration and investigation.

Future works may include additional ablation studies. Firstly, we can further evaluate whether the primary factor affecting \ac{fid} and \ac{map} performances arises from the overall quality of the generated image, including the background, or solely from that of the targets. Secondly, we can explore the influence of style transfer by not providing to the generative model any reference picture, relying solely on its pretrained knowledge. Finally, a comprehensive investigation into the effectiveness of the precision-recall filtering strategy, and the implications of employing a flawed ground truth during the training, demands further studies.

In conclusion, our method offers a novel and efficient solution to tackle the data scarcity challenge in object detection without the need for manual labeling of data, with the help of a pretrained generative model. In this way, we pave the way for further exploration of generative models applications in this field. This can be useful in specific domains where the gathering and labeling of data can be problematic or costly, such as, among others, medicine and molecular biology, atmospheric monitoring, material science, space imaging, and non-human-friendly environments.

\section{Acknowledgments}
M.P. acknowledges the contribution of the MAREA project funded by the Tuscany Region. S.M. acknowledges financial support from PNRR MUR project PE0000023-NQSTI. C.G. acknowledges the contribution of the National Recovery and Resilience Plan, Mission 4 Component 2 – Investment 1.4 – CN\_00000013 ``CENTRO NAZIONALE HPC, BIG DATA E QUANTUM COMPUTING'', spoke 6. C.G. and M.P. are members of the INdAM research group GNCS. The INdAM-GNCS support is gratefully acknowledged. F.C. acknowledges financial support by the European Commission's Horizon Europe Framework Programme under the Research and Innovation Action GA n. 101070546-MUQUABIS, by the European Union's Horizon 2020 research and innovation programme under FET-OPEN GA n. 828946-PATHOS, by the European Defence Agency under the project Q-LAMPS Contract No. B PRJ-RT-989, and by the MUR Progetti di Ricerca di Rilevante Interesse Nazionale (PRIN) Bando 2022 - project n. 20227HSE83 (ThAI-MIA) funded by the European Union--Next Generation EU.

\section*{Appendix A}
The standard ResNet-50 model starts with an initial convolutional layer with 64 $7\times 7$ filters, followed by a $3\times 3$ max pooling layer. The main body of ResNet-50 consists of a series of residual blocks, each functioning as a bottleneck. Each bottleneck comprises a $1\times 1$ convolutional layer to reduce dimensions, a $3\times 3$ convolutional layer to extract spatial features, and another $1\times 1$ convolutional layer to restore the original dimensions, totaling 3 convolutional layers per block.
The architecture contains 16 bottleneck blocks arranged in four stages, for a total of 48 convolutional layers: the first stage has 3 blocks with 256 channels, the second stage has 4 blocks with 512 channels, the third stage has 6 blocks with 1024 channels, and the final stage has 3 blocks with 2048 channels.
The network concludes with a global average pooling layer followed by a fully connected layer for classification.

In the Faster R-CNN framework, the ResNet50 backbone's final fully connected layer is omitted. Instead, the outputs of the four intermediate residual blocks, before each upsampling and before the final fully connected layer, are used as inputs for a \ac{fpn}~\cite{lin2017feature}. The \ac{fpn} is composed of two convolutional layers for each input: a lateral connection with $1\times 1$ filters to reduce the number of channels to 256, and a top-down pathway with a $3\times 3$ convolutional layer with 256 filters. The lateral connections process each input map, while the top-down pathway refines the feature maps, which are combined with the lateral outputs via summation. The final feature map is processed with another $3\times 3$ convolutional layer and a $1\times 1$ layer that proposes the regions of interest.

Two-stage detectors like Faster R-CNN operate in two phases: first generating region proposals (potential bounding boxes) and then refining and classifying these regions. By contrast, \ac{fcos} is a single-stage detector directly predicting bounding boxes and class scores from feature maps in a single pass.
\ac{fcos} eliminates the need for anchor boxes and proposals, streamlining the detection process. It uses a fully convolutional approach to predict bounding box coordinates for each pixel in the feature map. Key components of \ac{fcos} include a \ac{fpn} that generates feature maps at multiple scales with strides of 8, 16, 32, 64, and 128. Each level of the feature pyramid passes through shared heads comprising three parallel branches: classification, bounding box regression, and centerness. The classification branch outputs class scores, while the regression branch predicts the offsets for the bounding boxes. The centerness branch predicts a score indicating how close a pixel is to the center of an object, helping to down-weight low-quality bounding boxes far from object centers.

The MobileNetV3-Large architecture consists of a series of convolutional layers and bottleneck blocks designed for efficient mobile performance. It begins with a standard 3x3 convolutional layer with 16 output channels, followed by 16 bottleneck blocks with varying expansion sizes and kernel sizes, alternating between 3x3 and 5x5 kernels. Some of these blocks incorporate the Squeeze-and-Excite module, which enhances the network's representational power by adaptively recalibrating channel-wise feature responses through global average pooling and fully connected layers, culminating in a sigmoid activation. Following the bottleneck layers, the architecture includes a 1x1 convolutional layer with 960 output channels, a 7x7 pooling layer, and another 1x1 convolutional layer without batch normalization, expanding to 1280 output channels.

\vspace{1cm}
\bibliographystyle{unsrt}
\bibliography{biblio}
\end{document}